\documentclass{article}
\usepackage{spconf,amsmath,graphicx}
\usepackage[dvipsnames]{xcolor}
\usepackage{multirow}
\newcommand{\tabincell}[2]{\begin{tabular}{@{}#1@{}}#2\end{tabular}}

\title{Knowledge Augmented BERT Mutual Network in Multi-turn Spoken Dialogues}
\name{Ting-Wei Wu$^1$, Biing-Hwang Juang$^1$}
\address{
  $^1$Georgia Institute of Technology\\
  Department of Electrical and Computer Engineering\\
  \texttt{waynewu@gatech.edu, juang@ece.gatech.edu}}

\begin{document}
%
\maketitle
\begin{abstract}

Modern spoken language understanding (SLU) systems rely on sophisticated semantic notions revealed in single utterances to detect intents and slots. However, they lack the capability of modeling multi-turn dynamics within a dialogue particularly in long-term slot contexts. Without external knowledge, depending on limited linguistic legitimacy within a word sequence may overlook deep semantic information across dialogue turns. In this paper, we propose to equip a BERT-based joint model with a knowledge attention module to mutually leverage dialogue contexts between two SLU tasks. A gating mechanism is further utilized to filter out irrelevant knowledge triples and to circumvent distracting comprehension. Experimental results in two complicated multi-turn dialogue datasets have demonstrate by mutually modeling two SLU tasks with filtered knowledge and dialogue contexts, our approach has considerable improvements compared with several competitive baselines.

\end{abstract}
\section{Introduction}
\label{sec:intro}


Recent advances of spoken language understanding (SLU) modules prompt the success of task oriented dialogue systems, in transforming utterances into structured and meaningful semantic representations for dialogue management \cite{weld2021survey, su21_interspeech}. It mainly detects associated dialogue acts or intents and extracts key slot information as so-called \textit{`semantic frames'} \cite{abbeduto_1983}, shown in Table \ref{tab:dialog_ex}. Some knowledge triples in a knowledge base may be related to specific keywords in the dialogue which may accelerate the understanding process.

In early attempts of SLU tasks, isolated utterances in dissected dialogues were analyzed separately for user intents and semantic slots \cite{raymond07_interspeech, liu2017}. However, such ambivalent treatment hinders the transitions of shared knowledge for each supervised signal. Models that maximize the joint distribution likelihood were then proposed to amend the gap \cite{liu2016attentionbased, wang2018bimodel, wu-etal-2021-spoken}, 
with most studying the benefits of intent information for the later slot filling task. Some works also predicted multiple intents \cite{qin2019stackpropagation, Rash:19, wu-etal-2021-label}.
While driven by large pretrained corpus, these methods still fall short of employing complete dynamic interactions within dialogues. In contrast, humans can naturally adopt history contexts to identify intentions with their background knowledge. Some works have integrated previous dialogue contexts for more robust SLU \cite{wu21d_interspeech, knowledge_wang20, gupta2019casanlu}. 

\begin{table}[t]
  \small
  \caption{Snippet of a single turn within a dialogue with corresponding dialogue acts, slots and knowledge samples related to \textbf{keywords} in the utterance.}
  \label{tab:dialog_ex}
  \vspace{5pt}
  \centering
  \begin{tabular}{|l|l|}
    \hline
    \textbf{Speaker} & \textbf{Utterance} \\
    \hline\hline
    \textbf{1. User} & \tabincell{l}{Is there something that's \\ maybe a good intelligent \textbf{comedy}?} \\
    \hline
    \textbf{Act \& Slots}:  & \textit{Request\ (genre: \textbf{comedy})} \\
    \hline
    \textbf{Knowledge}: & \textit{(\textbf{comedy}; related to; comic)}\\
                        & \textit{(\textbf{comedy}; is a; drama)}\\
    \hline\hline
    \textbf{2. System} & \tabincell{l}{Whiskey Tango \textbf{Foxtrot} is the only Adult \\ comedy I see playing in your \textbf{area}. \\ Would you like to try that?} \\
    \hline
    \textbf{Act \& Slots}:  & \textit{Inform\ (movie: Whiskey Tango \textbf{Foxtrot})} \\
                            & \textit{Inform\ (genre: Adult comedy)} \\
                            & \textit{Inform\ (distance constraints: in your \textbf{area})} \\
                            & \textit{Confirm\_question} \\
    \hline                        
    \textbf{Knowledge}: & \textit{(\textbf{foxtrot}; related to; dance)}\\
                        & \textit{(\textbf{area}; is a; region)}\\
                        
    \hline
  \end{tabular}
\end{table}

\begin{figure*}[t]
  \centering
  \includegraphics[width=0.9\linewidth]{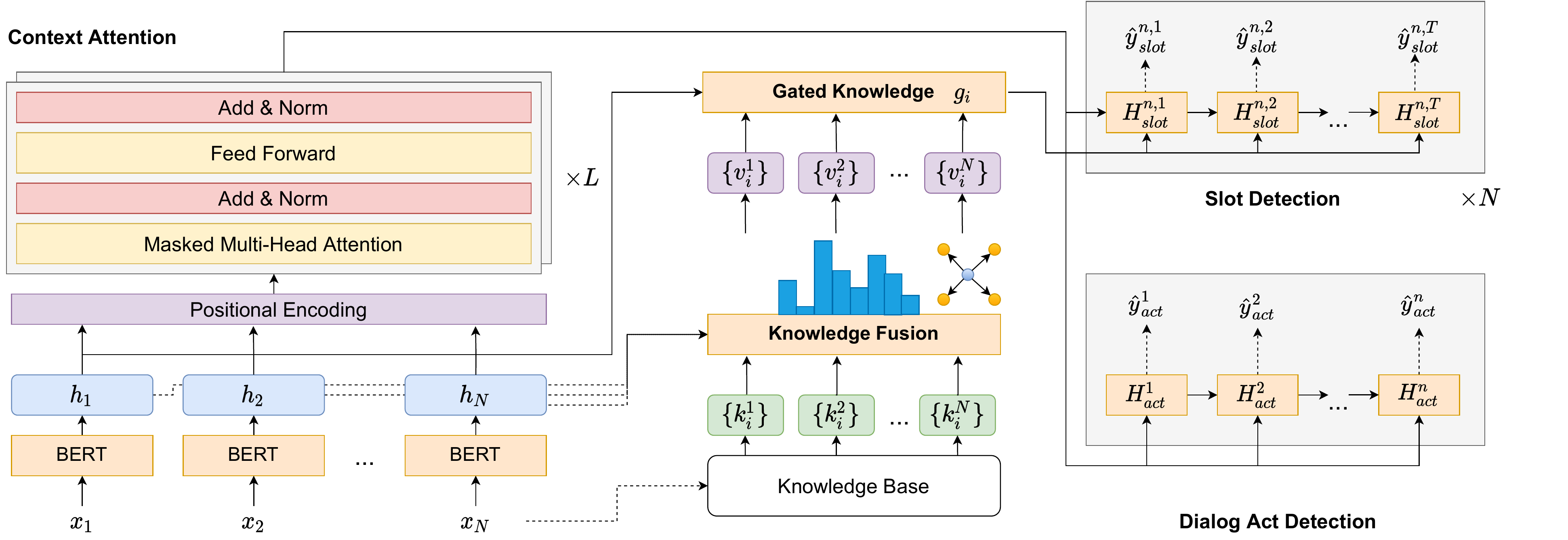}
  \caption{Illustration of our proposed framework for joint dialogue act detection and slot filling in multi-turn dialogs.}
  \label{fig:model}
\end{figure*}

Nevertheless, inadequacy of considering external knowledge may limit the machine to fully digest contexts and set constraints of comprehension boundaries. Much efforts have pushed forward the progress in knowledge grounded dialogue generation \cite{wang_generation21, zhao2020knowledgegrounded, zheng-etal-2021-knowledge}, where relevant documents or a knowledge base auxiliarily guide the language autoregressive progress. Term-level denoising \cite{zheng-etal-2021-knowledge} or filtering techniques \cite{wang_generation21} refine the adopted knowledge for better semantic considerations. Therefore, utilizing the correlation between language and knowledge is also imperative to some extent diminish ambiguity in dialogue context understanding, which recent SLU works often neglect. \cite{knowledge_wang20} has proposed to adopt knowledge attention for joint tasks. However, it adopts a single LSTM layer to couple all knowledge without filtering and contexts, which cannot model complex interactions well.

To solve above concerns, we propose a new \textbf{K}nowledge \textbf{A}ugmented \textbf{BE}RT \textbf{M}utual Network (KABEM) to effectively incorporate dialogue history and external knowledge in joint SLU tasks. Encoded knowledge is further gated to abate useless information redundancy. 
We then respectively induce dialogue contexts and knowledge to mutually predict intents and slots coherently with two LSTM decoders.
Experiment results have shown superior performance of our methods in manipulating contexts and knowledge for joint tasks and beat all competitive baselines. Our contributions are as follows:

\noindent 1.  We propose KABEM to incorporate external knowledge and previous dialogue history for joint multiple dialogue act and slot filling detection, where previous SLU works usually isolate the utterances without knowledge grounded. \\
2. We demonstrate the effectiveness of knowledge attention and the gating mechanism to reinforce the knowledge transitions between dialogue act and slot detection. \\
3. Experimental results show that our model achieves superior performances over several competitive baselines with more comprehensive knowledge consideration.

\section{Methodology}
\label{sec:method}

\subsection{Problem Statement}

In a dialogue $X = \{x_1, \dots, x_N\}$ of total $N$ user utterances and system responses, we would like to detect one or more dialogue acts $A$ and slots $S$ for each $x_n$. We denote the dialogue history $C_n = \{x_1, \dots, x_{n-1}\}$ and associated knowledge $K_n = \phi(K_G, x_n)$ for the current utterance $x_n$. $K_G$ is an external large knowledge base with knowledge triples and $\phi(\cdot)$ is the filter function. In essence, the joint probability distributions of predicting dialogue acts and slot labels are given as $A, S = arg \max P(A, S|x_n, C_n, K_n)$. For an utterance of $T$ words $x_n = \{w^n_1, w^n_2, \dots, w^n_T \}$, we will finally obtain a corresponding dialogue act set $\{a_i\}$ and a sequence of slot tags $\{s^n_1, s^n_2, \dots, s^n_T\}$.

\subsection{Context Attention}

To fully leverage the dialogue context information, we propose to encode the dialogue at token and turn levels respectively. At token level, we adopt BERT \cite{devlin2019bert}, a powerful NLP representation model, to extract semantic representations. For each utterance $x_n$ in a dialogue $X$, we encode it with BERT and obtain token-level representations $H = \{h_1, h_2, \dots, h_N \}$ from [CLS] tokens for $N$ utterances.

At turn level, to better capture semantic flows within a dialogue, we further encode $H$ with a context-aware unidirectional transformer encoder \cite{attention17}, which contains a stack of $L$ layers with each layer of a masked multi-head self-attention sublayer (MHA) and a point-wise feed forward network (FFN) with residual mechanism and layer normalization. We will send $H \in R^{N \times H_b}$ as the first layer input $C^1$ and iteratively encode with two sublayers in Eq. \ref{eq:all}. For each layer, it will first project the input $C$ with weight matrices: $W^Q, W^K, W^V \in R^{H_b \times H_a}$ to be $C^Q = C W^Q$, $C^K = C W^K$, $C^V = C W^V$.
Then each of them will be separated into $h$ heads, with each head $i$ to be $C_i \in R^{N \times (H_a/h)}$, $H_a$ is the hidden size for the attention module and $H_b$ is BERT hidden size. These $C_i$ will be sent into a self-attention and a feed forward layer in Eq.\ref{eq:att} and Eq.\ref{eq:ffn}. Finally, we will obtain the final contextual dialogue representations $C^{L}$.
\begin{align}
    \label{eq:all}
    C^l = FFN(MH&A(C^{l-1}, C^{l-1}, C^{l-1})) \\
    \label{eq:att}
    MHA(C^Q_i,C^K_i,C^V_i) &= softmax(\frac{C^Q_i(C^K_i)^T}{\sqrt{H_b}})C^V_i \\
    \label{eq:ffn}
    FFN(x) = max(&0,\ xW_1 + b_1)W_2 + b_2
\end{align}

\subsection{Knowledge Fusion}
\label{sec:kf}

To simulate the human awareness of coherently relating current contexts to background knowledge, the knowledge subgraph $k^n_i$ corresponding to the $i$-th word $w^n_i$ in $n$-th utterance $x_n$ is retrieved from the knowledge base $K_G$ using similar word matching. Each $k^n_i$ is a collection of multiple related triples $\gamma = \{h,r,t\}$, as head entity, relation, and tail entity. For each word, we then adopt an attention mechanism to dynamically filter irrelevant knowledge triples based on word contexts and obtain the knowledge-aware vector $v^n_i$.
\begin{align}
    v^n_i = &\sum_{j=1}^M \alpha_{ij} [r_{ij};t_{ij}] \\
    \alpha_{ij} = exp(&\beta_{ij}) / \sum_{m=1}^M exp(\beta_{im}) \\
    \beta_{ij} = (h^n_i W^H) (&tanh(r_{ij} W^R + t_{ij} W^T))^T
\end{align}
$r_{ij}$, $t_{ij}$ are relation and tail entity vectors. $W^H, W^R, W^T$ are learnable matrices during training. $M$ is the number of knowledge triples. $[;]$ is the concatenation of two vectors. Given the token-level representations for each word $h^n_i$ in the utterance $x_n$, attention weights are assigned to reveal the relevance of each knowledge triple under current contexts.

\subsection{Gated Knowledge}

Knowledge triples are mostly associated with name entities, where stochastic numbers or dates mentioned in utterances may not be relevant. We instead replace the triple vectors as zero vectors to represent agnosticism of knowledge, which will nonetheless introduce redundant noises. Therefore, we propose a gated mechanism for each word $h^n_i$ to regulate the degree of knowledge $v^n_i$ induced for downstream tasks and prevent information from overloading.
\begin{align}
    h^{n_i'} = g_i \cdot h^{n_i} + (1-g_i) \cdot v^{n_i} \\
    g_i = \sigma(W_i [h^{n_i};v^{n_i}] + b_i)
\end{align}
Information from word hidden states and corresponding knowledge is introduced in a trainable fully-connected layer with a sigmoid layer to produce a knowledge gated score. Then the network will balance the degree of knowledge influencing the decoding outputs.

\subsection{Semantic Decoder}

After obtaining the knowledge-enriched representations $H_K = \{h^{n_i'}\}$ along with contextual dialogue representations $C^{L}$, we adopt a BiLSTM for slot filling and a LSTM to detect multiple dialogue acts mutually. It will allow information to dynamically flow between two networks for understanding.
\begin{align}
    H_{slot} &= BiLSTM(H_K, C^L) \\
    H_{act} &= LSTM(C^L)
\end{align}
Knowledge-enriched vectors $H_K$ will be the inputs of BiLSTM with $C^L$ as initial hidden states, where contexts will assist the slot prediction at each knowledge-enhanced time step.
At the same time, we also input dialogue contexts $C^L$ only to another unidirectional LSTM for dialogue act detection since our context attention module is shared and has learned $H_K$ information implicitly. Finally, we can generate logits $\hat{y}_{act} = \sigma(H_{act}W_{act})$ by transforming $H_{act}$ with 
$W_{act} \in R^{H_L \times |\mathcal{Y}^a|}$
and a sigmoid function $\sigma$. $H_L$ is LSTM hidden size and $|\mathcal{Y}^a|$ is the size of dialogue act set. Likewise, we compute $\hat{y}_{slot} = softmax(H_{slot}W_{slot})$. Total loss will be the combination between the binary cross entropy loss based on $\hat{y}_{act}$ and the cross entropy loss based on $\hat{y}_{slot}$.

\begin{table*}[t]

  \caption{Experimental Results on several SLU models and ablation study of KABEM (\%). ID (Acc) indicates the dialogue act detection accuracy when all acts are predicted correctly. SL (F1) indicates the slot filling F1 score.}
  \label{tab:main}
  \small
  \vspace{5pt}
  \centering
  \begin{tabular}{|l||c|c|c|c|c|c|c|c|c|c|}
    \hline
    \textbf{Dataset} & \multicolumn{6}{c}{\textbf{MDC}} & \multicolumn{4}{|c|}{\textbf{SGD}} \\
    \hline
    \textbf{Domain} & \multicolumn{2}{c}{\textbf{Movie}}
                    & \multicolumn{2}{|c}{\textbf{Restaurant}}
                    & \multicolumn{2}{|c}{\textbf{Taxi}}
                    & \multicolumn{2}{|c}{\textbf{Restaurant}}
                    & \multicolumn{2}{|c|}{\textbf{Flights}}\\
    \hline
    \textbf{Model}  & ID (Acc) & SL (F1) & ID (Acc) & SL (F1) & ID (Acc) & SL (F1) & ID (Acc) & SL (F1) & ID (Acc) & SL (F1)\\
    \hline\hline
    MID-SF ~\cite{Rash:19}               & 76.56 & 67.56 & 77.35 & 65.77 & 85.03 & 70.03 & 74.26 & 81.38 & 84.74 & 84.48\\
    ECA ~\cite{chauhan20}                & 77.10 & 69.72 & 77.56 & 66.85 & 86.61 & 71.28 & 87.98 & 84.87 & 95.16 & 87.91\\
    KASLUM \cite{knowledge_wang20}       & 81.86 & 73.32 & 80.76 & 68.36 & 88.31 & 74.07 & 86.81 & 87.82 & 92.87 & 90.05\\
    CASA ~\cite{gupta2019casanlu}        & 84.22 & 79.59 & 83.17 & 74.89 & 90.00 & 78.54 & 92.54 & 94.20 & 95.00 & 91.79\\
    \hline\hline
    KABEM$_{AF}$ \cite{wang_generation21}     & 85.25 & 79.46 & 83.27 & 74.89 & 90.05 & \textbf{79.59} & 96.84 & 94.61 & 97.17 & 91.14\\
    KABEM                                & 85.63 & \textbf{80.03} & \textbf{83.69} & \textbf{75.36} & \textbf{90.95} & 79.18 & \textbf{97.70} & \textbf{96.63} & \textbf{98.10} & \textbf{94.02}\\
    \hline\hline
    \; w/o KG                           & \textbf{86.01} & 79.92 & 83.53 & 74.76 & 90.56 & 78.29 & 97.53 & 94.83 & 97.73 & 92.23\\
    \; w/o CA                           & 84.87 & 79.79 & 81.33 & 74.68 & 89.00 & 78.50 & 95.88 & 94.36 & 97.17 & 91.94\\
    \; w/o LSTM                         & 84.57 & 79.14 & 82.70 & 74.35 & 89.65 & 79.00 & 90.96 & 93.64 & 94.80 & 91.33\\
    
    \hline
  \end{tabular}
\end{table*}


\section{Experiments}
\label{sec:experiments}

\subsection{Experimental setup}

We evaluate our proposed framework on two large-scale dialogue datasets, i.e. Microsoft Dialogue Challenge dataset (MDC) \cite{li2018microsoft} and Schema-Guided Dialogue dataset (SGD) \cite{rastogi2019towards}. \textbf{MDC} contains human-annotated conversations in three domains (movie, restaurant, taxi) with total 11 dialogue acts and 50 slots. 
\textbf{SGD} entails dialogues over 20 domains ranging from travel, weather to banks etc. It has more structured annotations with total 18 dialogue acts and 89 slots. 
We randomly select 1k dialogues for each domain in \textbf{MDC} and the restaurant domain from \textbf{SGD} to compare that in MDC and a very different domain (flights) for total 5k dialogues in 7:3 training and testing ratio. 
Each utterance is labeled with one or more dialogue acts and several slots.

We compare our models with several competitive baselines which sequentially include more semantic features: 
    \noindent \textbf{MID-SF}~\cite{Rash:19} which first considers multi-intent detection with slot filling tasks with BiLSTMs.
    \noindent \textbf{ECA}~\cite{chauhan20} which encodes the dialogue context with a LSTM encoder for joint tasks.
    \noindent \textbf{KASLUM}~\cite{knowledge_wang20} which extracts knowledge from a knowledge base and includes dialogue history for joint tasks.
    \noindent \textbf{CASA}~\cite{gupta2019casanlu} which encodes the context with DiSAN sentence2token and we replace BERT encoder to demonstrate its contributions.
    \noindent \textbf{KABEM$_{AF}$}~\cite{wang_generation21} we replace only Knowledge Fusion part in KABEM ($\S$ \ref{sec:kf}) with the attention-based filter (AF) in \cite{wang_generation21} to compare different knowledge attention.


We adopt the pretrained $BERT_{base}$ \cite{devlin2019bert} as our utterance encoder. 
Context attention transformer has $L=6$-layer attention blocks with 768 head size and 4 attention heads. 
The max sequence length is 60. 
We use simple string matching of words to extract relevant knowledge triples from the ConceptNet. 
Then, TransE \cite{transe13} is adopted to represent head, relation and tail as 100-dim vectors.
We retrieve 5 most related knowledge from each word based on weights assigned on the edges. Both LSTMs have 256 hidden units. We use the batch size of 4 dialogues for MDC and 2 for SGD. 
In all training, we use Adam optimizer with learning rate as 5e-5. The best performance on validation set is obtained after training 60 epochs on each model. For metrics, we report the dialog act accuracy and slot filling F1 score. Here we only consider a true positive when all BIO values for a slot is correct and forfeit `O' tags.

\section{Results and analysis}
\label{sec:results}

\subsection{Main results}

Table. \ref{tab:main} shows our main results on the joint task performances of several advanced neural network based frameworks. 
MID-SF with only LSTMs has relatively inferior performances on both datasets especially in SGD. 
ECA with dialogue contexts enhanced has much greater increase in SGD than in MDC and further knowledge induction gives 3.5 \% increase in KASLUM. 
Leveraging BERT-based encoder seems to substantially increase semantic visibility in CASA and KABEM. 
Eventually, KABEM$_{AF}$ and KABEM beat all baselines both in MDC and substantially in SGD, while our knowledge fusion module incorporates external knowledge and dialogue contexts more efficiently.

To better estimate the effectiveness of each module of KABEM, we conduct ablation experiments following in Table. \ref{tab:main}. 
We sequentially ablate each component from KABEM to observe the performance drops. By removing knowledge attention with gating (KG), we see more obvious reduction in slot filling tasks denoting the necessity of external knowledge. By substituting a unidirectional LSTM on top of BERT for our context attention module (CA), we obtain poorer performance in dialogue act detection instead. Finally, we see dialogue contexts are more crucial in SGD where drop seems significant by removing all context fusion modules. Overall, we observe dialogue act detection relies more on contexts while slot filling tasks may concentrate on inter-utterance relations where external knowledge benefits more instead.

\subsection{Knowledge attention}

In Table. \ref{tab:kg}, we visualize the extracted knowledge and their weights corresponding to three important keywords for semantic detection in the utterance. Here, the word `cheap' is super related to `affordable' which helps identifying the slot `pricing'. Our model also leverages the fact of `yesterday' and `tomorrow' to identify a `date' slot. Eventually, knowledge related to `city' assists the city identification for `Seattle', especially beneficial when model has never seen `Seattle' in the training data. To notice, numbers or time are not valid entities inside the knowledge base, where equal weights are assigned to each zero vector and our gating mechanism will circumvent from using it for prediction.

\begin{table}[t]
  \caption{A utterance example of utilizing knowledge for joint task prediction. Knowledge (Relation, Tail)  related to three \textbf{keywords} as head are presented with their attention weights. `rel' represents `related to' and `ant' represents `antonym'. }
  \label{tab:kg}
  \footnotesize
  \vspace{5pt}
  \centering
  \begin{tabular}{|l|l|l|}
    \hline
    \multicolumn{3}{|c|}{\textbf{Utterance Example}} \\
    \hline\hline
    \textbf{Utterance} & \multicolumn{2}{|c|}{\tabincell{l}{I need a \textbf{cheap} food place for \\ 3 people \textbf{tomorrow} at 1pm in \textbf{Seattle}.}} \\
    \hline
    \textbf{Dialog acts} & \multicolumn{2}{|c|}{Request} \\
    \hline
    \textbf{Slots} & \multicolumn{2}{|c|}{\tabincell{l}{O O O \textbf{B-pricing} O O O B-numberofpeople \\ O \textbf{B-date} O B-starttime I-starttime O \textbf{B-city}}} \\
    \hline
    \multicolumn{3}{|c|}{\textbf{Knowledge}}\\
    \hline
    \textbf{cheap} & \textbf{tomorrow} & \textbf{Seattle} \\
    \hline
    rel, affordable (0.99) & rel, later\_on (5e-2)        & rel, city\_usa (2e-2)\\
    rel, chintzy    (3e-7) & rel, morrow (7e-3)           & rel, washington (1e-4)\\
    rel, chinchy    (2e-9) & is a, future (9e-7)           & rel, emerald\_city (9e-2)\\
    rel, twopenny   (5e-5) & is a, day (4e-6)              & part of, wa (0.87) \\
    rel, gimcrack   (8e-6) & ant, yesterday (0.9)         & is a city\_wa (8e-3)\\
    \hline
  \end{tabular}
\end{table}

\section{Conclusion}
\label{sec:conclusion}

In this paper, we propose a novel BERT-based integrated network to both consider dialogue history and external knowledge in joint SLU tasks. The model is capable of selecting relevant knowledge triples and adopts the attention mechanism to acquire useful knowledge representation. Fused information is then mutually induced between the prediction of dialogue acts and slots. The effectiveness of our proposed model is verified in two multi-turn dialogue datasets and knowledge fusion vectors could be easily applied to downstream dialogue state tracking or management tasks.

\bibliographystyle{IEEEbib}
\bibliography{Template}

\end{document}